\documentclass[runningheads]{llncs}
\usepackage{graphicx}
\usepackage[utf8]{inputenc}
\usepackage{caption}
\usepackage{amssymb}
\usepackage{mathtools}
\usepackage{hyperref}
\usepackage{todonotes}

\usepackage{booktabs}
\usepackage{multirow}
\usepackage{adjustbox}
\usepackage{colortbl}

\usepackage{tikz}

\newcommand\copyrightnotice{%
    \begin{tikzpicture}[remember picture,overlay]
    \node[anchor=south,yshift=10pt] at (current page.south) {\fbox{\parbox{\dimexpr\textwidth-\fboxsep-\fboxrule\relax}{
    \footnotesize \textcopyright\ Springer, 2021. The final authenticated version is available online at:
\href{YOUR_DOI}{https://doi.org/10.1007/978-3-030-91431-8\_46}
    }}};
    \end{tikzpicture}%
}

\begin{document}

\title{LogLAB: Attention-Based Labeling of Log Data Anomalies via Weak Supervision}
\titlerunning{LogLAB: Attention Based Labeling of Log Data Anomalies}

\author{Anonymous}
\institute{Anonymous}

\author{Thorsten Wittkopp \and Philipp Wiesner \and Dominik Scheinert \and Alexander Acker}
\authorrunning{T. Wittkopp, et.al.}

\institute{Technische Universität Berlin\\
DOS, TU-Berlin, Germany\\
\email{\{t.wittkopp, wiesner, dominik.scheinert, alexander.acker\}@tu-berlin.de}}

\maketitle              

\begin{abstract}
With increasing scale and complexity of cloud operations, automated detection of anomalies in monitoring data such as logs will be an essential part of managing future IT infrastructures.
However, many methods based on artificial intelligence, such as supervised deep learning models, require large amounts of labeled training data to perform well.
In practice, this data is rarely available because labeling log data is expensive, time-consuming, and requires a deep understanding of the underlying system.
We present LogLAB, a novel modeling approach for automated labeling of log messages without requiring manual work by experts.
Our method relies on estimated failure time windows provided by monitoring systems to produce precise labeled datasets in retrospect.
It is based on the attention mechanism and uses a custom objective function for weak supervision deep learning techniques that accounts for imbalanced data.
Our evaluation shows that LogLAB consistently outperforms nine benchmark approaches across three different datasets and maintains an F1-score of more than 0.98 even at large failure time windows.

\end{abstract}

\section{Introduction}
\copyrightnotice
As more and more companies outsource their IT services to the cloud, the number of servers and interconnected devices is continuously increasing. 
In the meantime, modern abstraction layers are driving the creation of large multilayered systems while adding technical complexity under the hood. 
This aggravates the operation and maintenance of systems and services and, therefore, poses new challenges for cloud operators. 
To maintain control over complexity, monitoring becomes an integral part of cloud infrastructure operations.
However, in today's systems the amount of monitoring data is often growing to an extent that cannot be analyzed manually.

The area of artificial intelligence for IT operations (AIOps) is intended to support cloud operators to ensure operational efficiency as well as dependability and serviceability~\cite{gulenko2020ai}.
A core component of any AIOps system is the detection of anomalies in monitoring data such as metrics, logs, or traces.
Log data are one of the most important resources for troubleshooting because they record events during the execution of service applications.
However, even though most types of log messages come with a severity level, these do not necessarily reflect the status of the overall system.
Therefore, recent research utilizes deep learning models to analyze log data and perform anomaly detection~\cite{du2017deeplog,zhang2019robust,nedelkoski2020self}. 
One of the main obstacles in log anomaly detection is the lack of labeled log data~\cite{wittkopp2020decentralized}.
Labeling data is costly and time-consuming, as experts need to analyze every single log message and investigate which messages reflect their corresponding errors.
Since supervised models that train on large volumes of labeled data show significant performance in log anomaly detection~\cite{zhang2019robust,yang2021semi}, it is important to automate the labeling process to gain a strong accelerator for log anomaly detection \cite{ratner2016data}.

To address this problem, we propose LogLAB, an attention-based model for binary labeling of anomalies in log data via weak supervision.
It relies only on rough estimates of when an error has occurred - information that can often be derived from other monitoring systems~\cite{sukhwani2017monitoring}.
Specifically, the contributions of this paper are:

\begin{itemize}
    \item A problem description for how to label anomalies in log data using monitoring information and weak supervision including a method solving this.
    \item A custom objective function for weak supervision deep learning techniques that takes class-imbalanced data into account.
    \item An extensive evaluation of ten different approaches solving the defined problem, including LogLAB and its implementation\footnote{https://github.com/dos-group/LogLAB}.
\end{itemize}

The remainder of this paper is structured as follows.
\autoref{sec:related-work} surveys the related work.
\autoref{sec:approach} provided a problem description and explains our approach LogLAB.
\autoref{sec:evaluation} evaluates LogLAB in comparison to nine other approaches.
\autoref{sec:conclusion} concludes the paper.

\section{Related Work}\label{sec:related-work}
We discuss works for text classification, anomaly detection and PU learning.

\textbf{Text-Based Classification.}
Many established methods are discussed in~\cite{kowsari2019text}. 
The PCA algorithm~\cite{jolliffe2005principal} is for instance often employed for dimensionality reduction right before the actual classification procedure. 
Random forests~\cite{ho1995random} are another technique and a suitable tool due to their ensemble learning design. 
Logistic regression~\cite{hosmer2013applied} belongs to the classic statistical methods~\cite{genkin2007large}.
Other publications utilize the Rocchio algorithm that is compared against kNN in~\cite{sowmya2016large}. 
In another work~\cite{selvi2017text}, the authors design a pipeline involving the Rocchio algorithm. 

\textbf{Log Anomaly Detection.}
The experience report for anomaly detection on system logs~\cite{he2016experience} discusses additional methods. 
Invariant Miners~\cite{lou2010mining} retrieve structured logs using log parsing, further group log messages according to log parameter relationships, and mine invariants from the groups to perform actual anomaly detection on logs.
Decision Trees~\cite{quinlan1986induction} are another solution often employed in classification problem scenarios.
SVMs are evaluated in~\cite{manevitz2001one} for document classification and anomaly detection.
The authors of~\cite{schapire2000boostexter} propose a boosting-based system and thus ensemble learning method that shows good performance. 
Deep Learning methods are also more and more used in the realm of log anomaly detection.
DeepLog~\cite{du2017deeplog} utilizes an LSTM and thus interprets a log as a sequences of templates to performs anomaly detection per log message.
More recent works~\cite{yang2021semi,yang2019nlsalog,wittkopp2021a2log} also make use of deep learning.

\textbf{PU Learning.}
A problem setting also discussed in other works.
For instance, the authors in~\cite{liu2002partially} utilize the EM algorithm together with naive Bayesian classification.
A more conservative variant of this method is proposed in~\cite{fusilier2015detecting} where the set of reliable negative instances is iteratively pruned using a binary classifier, which ultimately leads to improved final prediction results due to the few but high quality negative instances. 
An ensemble learning method for PU learning is proposed in~\cite{mordelet2014bagging}. The authors motivate bagging SVM, i.e. the aggregation of multiple SVM classifiers in order to answer sources of instability often encountered in PU learning situations.

\section{Automated Log Labeling}\label{sec:approach}

\subsection{Problem Description}\label{sec:towards}
Log messages can describe failures that occur during runtime, such as the crash of a service.
We refer to such log messages as 'abnormal'.
Modern monitoring solutions raise alerts when a system runs into an abnormality or outages occur by observing metrics, hardware component failures, workload deployment failures and other failure scenarios~\cite{sukhwani2017monitoring}. 
Therefore, we assume that in an IT operation center failure time windows of services and systems are roughly known.
We use this information in retrospect to identify and label abnormal log messages.
\vspace{-0.3cm}

\begin{figure}[htbp]
\centering
\includegraphics[width=0.8\columnwidth]{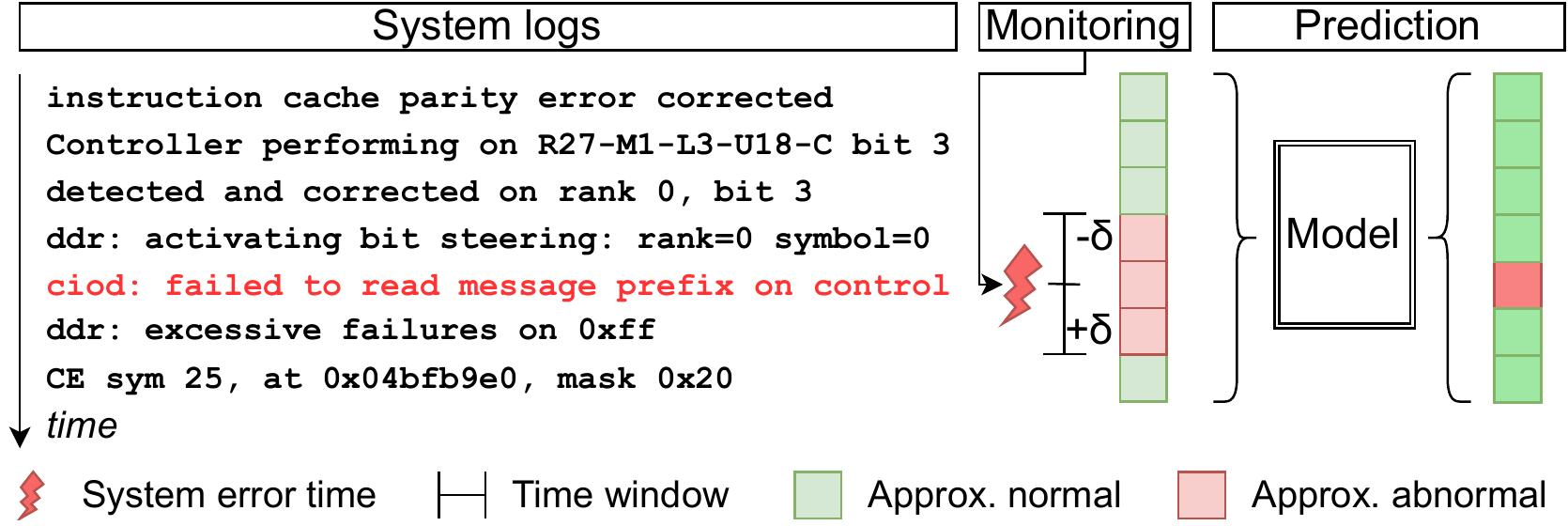}
\caption{We use rough estimates for failure times provided by monitoring systems in order to identify and label abnormal log messages via weak supervision.}
\label{fig:problem_description}
\end{figure}

\vspace{-0.3cm}
\autoref{fig:problem_description} provides an example for the described problem.
It displays the log of a system with one abnormal log event (colored in red).
We utilize monitoring information to estimate time windows of the length $2*\delta$ in which we suspect abnormal log events. 
The model's task is to identify the abnormal log messages in the time window and classify all others as normal.

We describe the log labeling as a weak supervision learning problem with inaccurate labels as defined by Zhou et. al~\cite{zhou2018brief}. Thereby, label inaccuracy stems from the imprecision of the failure time windows.
We assign inaccurate labels for all log events, depending on whether they are in the failure time windows or not. 
Further, we utilize PU learning \cite{liu2002partially,liu2003building} which is short for learning from positive and unlabeled data.
Thereby, the underlying log data is divided into two classes, positive $\mathcal{P}$ and unlabeled $\mathcal{U}$, where $\mathcal{U}$ consists of all log messages that occur in the aforementioned failure time windows and $\mathcal{P}$ of the remaining log messages. 

\subsection{LogLAB}
For the labeling of logs, we design a processing and modeling pipeline illustrated in~\autoref{fig:high_level_pipieline}. The individual steps are as follows:
\vspace{-0.5cm}

\begin{figure}[htbp]
\centering
\includegraphics[width=0.8\columnwidth]{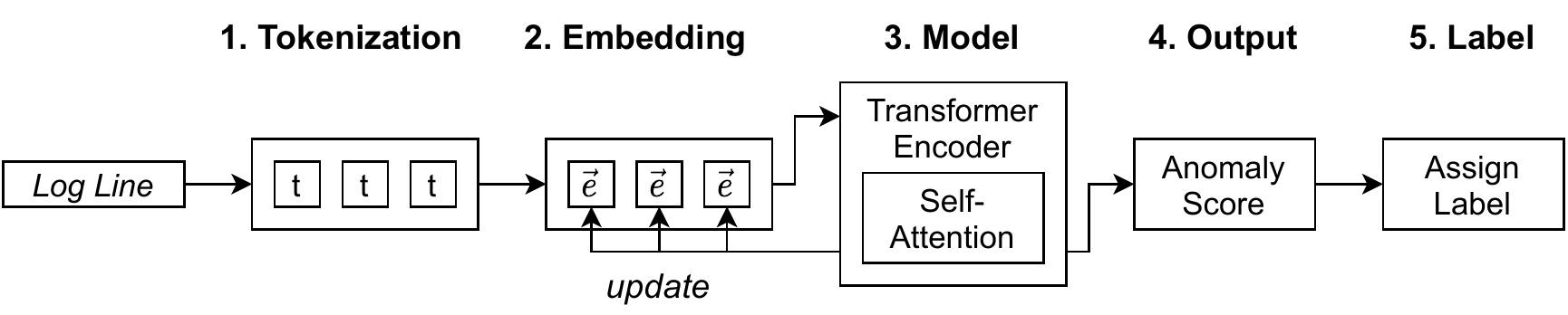}
\caption{High level log message labeling pipeline.}
\label{fig:high_level_pipieline}
\end{figure}

\vspace{-0.5cm}
First, we convert the content $c_i$ of each log message $l_i$ into a sequence of tokens $t_i$ by splitting on the symbols \texttt{.,:/} and whitespaces. 
Subsequently, we clean the resulting sequence of tokens by replacing certain tokens with placeholders. 
Thereby placeholder tokens for hexadecimal values \texttt{'[HEX]'} and any number greater or equal 10  \texttt{'[NUM]'} are introduced. Finally, we prefix the sequence of transformed tokens with a special token \texttt{'[CLS]'} which serves as a numerical summary of the whole log message. 
An exemplary log message: \texttt{time.c: Detected 3591.142 MHz}
is thus transformed into a sequence of tokens: \texttt{['[CLS]', 'time', 'c', 'Detected', '[NUM]', '[NUM]', 'MHz']}. 

Since these sequences can vary in length, we truncate them to a fixed size and pad smaller sequences with \texttt{'[PAD]'} tokens.
For each token $w_j$ of the token sequence $t_i$, an embedding $\vec{e}_{i}(j)$ is obtained. 
The truncated sequences of embeddings $\vec{e'_i}$ serves as the input for the model. 

The model computes an output embedding, for each input sequence $\vec{e'_i}$, which summarizes the log message by utilizing the embeddings of all tokens. 
This output embedding is encoded in the embedding of the \texttt{’[CLS]’} token which is also modified during training. 
For this purpose, we utilize the transformer architecture~\cite{DevlinCLT19} with additional self-attention~\cite{VaswaniSPUJGKP17}.
During the training process, the model is supposed to learn the meanings of the log messages, thereby getting an intuition of what is normal and abnormal.
Finally, this model outputs a vector (embedding) for each input sequence $\vec{e'_i}$. 
We denote the output of the model as $z_i=\Phi(\vec{e'_i};\Theta)$ and use it throughout the remaining steps.
Thereby the anomaly score is calculated by the length of the output vector $\lVert z_i \rVert$.
Anomaly scores close to $0$ represent normal log messages, whereby large vectors indicate an abnormal log message. 
The computed anomaly score is used to assign a label $\widehat{y_i}$ to the log message $l_i$, i.e. either normal or abnormal.

\subsection{Objective Function}
To label the log data, the model must be trained in a way that it is capable to handle the problem of weak supervision with inaccurate labels. 
Thus, the objective function must assign log anomaly scores to log messages that occur in class $\mathcal{P}$ and $\mathcal{U}$. 
Log messages that occur only in $\mathcal{U}$ are likely abnormal and must therefore have higher anomaly scores.
In addition, the loss function must be able to handle large amounts of incorrectly labeled log messages, since the class $\mathcal{U}$ can increase quickly for large $\delta$.
The objective function consist of two parts. 
The first part minimize the errors of samples from class $\mathcal{P}$, from which the calculated anomaly scores should be close to $0$. 
The second part of the objective must minimize the errors of samples from class $\mathcal{U}$, by pushing them away from $0$. 
The structure of the objective function is defined as $\frac{1}{m}\sum\limits_{i=1}^{m}((1-\tilde{y}_i)*a(z_i) + (\tilde{y}_i)*b(z_i)$, where $\tilde{y}_i$ is the inaccurate label, $z_i$ the output vector and $m$ the batch size.
The function '$a()$' becomes $0$ if the sample is from class $\mathcal{U}$, while the second function '$b()$' becomes $0$ if the sample is from class $\mathcal{P}$. 
For $a$ we choose $a(z_i) = \lVert z_i \rVert^2$ and for $b$ we choose $b(z_i) = \frac{q^2}{\lVert z_i \rVert}$ to minimize the error. Thereby $a$ calculates the squared error of the length of the output for samples from class $\mathcal{P}$. In contrast, we increase the error for all small anomaly scores when the log message is of class $\mathcal{U}$. Thereby $q$ is a numerator between 0 and 1 that represents the relation of the number of samples in $\mathcal{P}$ and $\mathcal{U}$. 
To ensure that $q$ is representing the relation of $\mathcal{P}$ and $\mathcal{U}$ and remains in the boundaries of 0 to 1, we model $q$ as a limited function $f(x) = \frac{x}{x+1}$, with $\lim \limits_{x \to \infty} f(x) = 1$, that is provided with the relation of $\mathcal{P}$ and $\mathcal{U}$: $q = f(\frac{|\mathcal{P}|}{|\mathcal{U}|}) = \frac{\frac{|\mathcal{P}|}{|\mathcal{U}|}}{(\frac{|\mathcal{P}|}{|\mathcal{U}|}+1)} = \frac{|\mathcal{P}|}{|\mathcal{P}|+|\mathcal{U}|}$.
Thus the total loss function can be expressed as: $\frac{1}{m}\sum\limits_{i=1}^{n}\Big((1-\tilde{y}_i)*\lVert z_i \rVert^2 + (\tilde{y}_i)*\frac{(\frac{|\mathcal{P}|}{|\mathcal{P}|+|\mathcal{U}|})^2}{\lVert z_i \rVert}\Big)$.

\section{Evaluation}\label{sec:evaluation}

To obtain a significant and wide benchmark, we compare LogLAB to several state of the art text-classification and anomaly detection approaches presented in a recent text-classification survey~\cite{kowsari2019text} as well as in an established survey for anomaly detection in system logs~\cite{he2016experience}.
Namely, we choose PCA, Invariant Miners, Deeplog, Decision Trees, Random Forests, SVMs, Logistic Regression, the Rocchio algorithm, and boosting approaches as benchmark methods. 
Thereby we measure the deviation from the ground truth $y_i$ and the calculated labels $\widehat{y_i}$.

\subsection{Experimental Setup}

We evaluate all methods on three labeled log datasets recorded at different large-scale computer systems\cite{oliner2007datasets}.
The \emph{BGL} dataset contains 4\,747\,963 log messages of which 7.3\,\% are abnormal and records a period of ~214 days, with on average 0.25 log messages per second.
We selected the first 5 M log messages from the \emph{Thunderbird} dataset of which 4.5\,\% are abnormal. They account for a period of ~9 days, with on average 6.4 log messages per second.
Again, we selected the first 5 M log messages from the \emph{Spirit} dataset of which 15.3\,\% are abnormal. They cover a period of ~48 days, with on average 1.2 log messages per second.

    
    

We create our evaluation datasets with inaccurate labels by including all abnormal log events as well as their surrounding events within a time window $2*\delta$ in $\mathcal{U}$; all remaining log events are in $\mathcal{P}$. 
Thereby we investigate the performance at three different time windows: $\pm 1000\,ms$ (2s), $\pm 5000\,ms$ (10s) and $\pm 15000\,ms$ (30s).
The amount of samples in $\mathcal{U}$ is changing for BGL: $0.39M$, $~0.44M$ and $~0.47M$,
Thunderbird: $1.42M$, $2.36M$ and $2.90M$ and 
Spirit: $1.00M$, $2.33M$ and $3.26M$ regarding the respective time window $\delta$.

Each sequence of tokens $t_i$ is truncated to have a length of 20 for \emph{Thunderbird}, 16 for \emph{Spirit}, and 12 for \emph{BGL}. 
The dimensionality $d$ of our embeddings is set to 128.
For the training of our LogLAB model, we use a hidden dimensionality of 256, a batch size of 1024, a total of 8 epochs, and a dropout rate of 10\%. 
We use the Adam optimizer with a learning rate of $10^{-4}$ and a weight decay of $5 \cdot 10^{-5}$. 

\subsection{Results}

\begin{adjustbox}{
center,label={table:results},caption={Evaluation results: F1-scores above 0.99 and 0.98 are highlighted in blue and cyan, respectively.},float=table}
	
\resizebox{0.92\textwidth}{!}{%
\begin{tabular}[h!]{p{1.6cm} p{1.5cm} p{1cm} p{1cm} p{1.2cm}  p{1cm} p{1.05cm} p{1cm} p{1cm} p{1cm} p{1cm} p{1cm} } 
	\toprule
 	& & \multicolumn{3}{c}{Learning $\mathcal{P}$} & \multicolumn{7}{c}{Learning $\mathcal{P}$ and $\mathcal{U}$} \\
 	\cmidrule(lr){3-5}
 	\cmidrule(lr){6-12}
	\rotatebox{90}{Dataset} & Metric & \rotatebox{45}{PCA} & \rotatebox{45}{Invariant Miners} & \rotatebox{45}{Deeplog} & \rotatebox{45}{Decision Tree} & \rotatebox{45}{Random Forest} & \rotatebox{45}{SVM} & \rotatebox{45}{Logistic Regr.} & \rotatebox{45}{Boost} & \rotatebox{45}{Rocchio} & \rotatebox{45}{\textbf{LogLAB}} \\ [0.5ex]
	
 	\midrule
	\multicolumn{12}{c}{$\delta=\pm 1000ms$} \\
	\midrule
 
 	BGL & F1-Score   & 0.5963 & 0.5102 & 0.7759 & \cellcolor{blue!20}0.9974 & \cellcolor{cyan!20}0.9830 & \cellcolor{cyan!20}0.9840 & \cellcolor{blue!20}0.9976 & \cellcolor{blue!20}0.9908 & 0.7096 & \cellcolor{blue!20}0.9977 \\ \midrule
 
 	TBird & F1-Score   & 0.3048 & 0.1824 & 0.0880 & 0.3242 & 0.3144 & 0.3235 & 0.3242 & {0.3361} & 0.3440 & \cellcolor{blue!20}0.9995 \\ \midrule
 
 	Spirit & F1-Score   & 0.8043 & 0.5807 & \cellcolor{blue!20}0.9926 & \cellcolor{blue!20}0.9967 & 0.9604 & \cellcolor{cyan!20}0.9857 & \cellcolor{blue!20}0.9962 & \cellcolor{blue!20}0.9968 & \cellcolor{blue!20}0.9971 & \cellcolor{blue!20}0.9997 \\ 
 	
 	\midrule
	\multicolumn{12}{c}{$\delta=\pm 5000ms$} \\
	\midrule
 
 	BGL & F1-Score   & 0.5930 & 0.5112 & 0.7755 & \cellcolor{cyan!20}0.9874 & 0.9646 & 0.9680 & \cellcolor{cyan!20}0.9875 & 0.9795 & 0.8054 & \cellcolor{blue!20}0.9949 \\ \midrule
 
 	TBird & F1-Score   & 0.3053 & 0.1936 & 0.0651 & 0.2669 & 0.2415 & 0.2439 & 0.2678 & 0.2869 & 0.3146 & \cellcolor{blue!20}0.9995 \\ \midrule
 
 	Spirit & F1-Score   & 0.7691 & 0.5740 & \cellcolor{blue!20}0.9929 & 0.6513 & 0.5453 & 0.5584 & 0.6560 & 0.5830 & \cellcolor{blue!20}0.9946 & \cellcolor{blue!20}0.9980 \\
 	
 	\midrule
	\multicolumn{12}{c}{$\delta=\pm 15000ms$} \\
	\midrule
 	BGL & F1-Score   & 0.5879 & 0.5130 & 0.7760  & 0.9753 & 0.9483 & 0.9523 & 0.9767 & 0.9762 & 0.7898 & \cellcolor{blue!20}0.9902 \\ \midrule
 
 	TBird & F1-Score   & 0.3025 & 0.1933 & 0.1113 & 0.1341 & 0.1248 & 0.1348 & 0.1350 & 0.1476 & 0.2241 & \cellcolor{blue!20}0.9995 \\ \midrule
 
 	Spirit & F1-Score   & 0.4958 & 0.4254 & 0.8236 & 0.4909 & 0.4735 & 0.4836 & 0.4917 & 0.4887 & 0.5192 & \cellcolor{cyan!20}0.9825 \\ 
 	\bottomrule
\end{tabular}
}
\end{adjustbox}

To compare LogLAB to our baselines, we assess the prediction performance $\tilde{y}_i \sim y_i$ in terms of F1-score metrics. The F1-scores are presented in \autoref{table:results}. 
As expected, with increasing $\delta$ and thus growing size of $\mathcal{U}$, the performance across all approaches tends to decrease.
For $\delta = \pm 1000\,ms$ this was apparently easy to achieve for most of the methods. An exception is the Thunderbird dataset, which is characterized by a large $\mathcal{U}$ class: No baseline manages to achieve an F1-score higher than 0.35, except LogLAB.
For $\delta = \pm 5000\,ms$ we notice that the performance degradation previously observed for most approaches on the Thunderbird dataset now also start to manifest on the Spirit dataset. 
The biggest gap in performance becomes evident at the largest observed time window of $\delta = \pm 15000\,ms$. 
For the dataset BGL, we notice a considerable drop in F1-scores of other approaches to 0.97, while LogLAB maintains its high performance. 

\section{Conclusion}\label{sec:conclusion}
This paper presents LogLAB, a novel model for labeling large amounts of log data, such that the usually required need of time-consuming manual labeling through experts is automated.
It relies only on rough estimates of failure time windows provided by monitoring systems to generate labeled datasets in retrospect.
LogLAB is based on the attention mechanism and uses a custom objective function for weak supervision deep learning techniques that accounts for imbalanced data and deals with inaccurate labels. We evaluated LogLAB on three different datasets in comparison to nine benchmark approaches.
LogLAB outperforms other approaches across all experiments and shows a performance of more than $0.98$ F1-score, even for large amount of inaccurate labels.
As further work we consider to enhance the labeling process by iteratively moving log messages from $\mathcal{U}$ to $\mathcal{P}$ during training, which have significantly lower anomaly scores, calculated by the model. Likewise, this method can be extended by adding other sources for estimating the time windows and therefore improve the training basis.

\bibliographystyle{splncs04}
\bibliography{bib}
\end{document}